\documentclass[sigconf]{acmart}

\usepackage{booktabs} % For formal tables 
\usepackage{times}
\usepackage{latexsym}

\usepackage[T1]{fontenc}
\usepackage[utf8]{inputenc}

\usepackage{microtype}
\usepackage{graphicx}
\usepackage{url}
\usepackage{latexsym}
\usepackage{amsmath}
\usepackage{amsfonts}
\usepackage{algorithmic}

\usepackage{graphicx}
\usepackage{textcomp}
\usepackage{xcolor}
\usepackage{tabularx}
\usepackage{pgfplots}
\usepackage[ruled,vlined]{algorithm2e}
\usepackage{dirtytalk}
\usepackage{booktabs}
\usepackage{caption}
\usepackage{subcaption}
\usepackage{wrapfig}
\usepackage{lipsum}
\usepackage{setspace}
\usepackage[title]{appendix}
\usepackage{longtable}
\usepackage{multirow}
\usepackage{csquotes}

\setcopyright{acmcopyright}
\copyrightyear{2023}
\acmYear{2023}
\setcopyright{rightsretained}
\acmConference[SAC '23]{The 38th ACM/SIGAPP Symposium on Applied Computing}{March 27-March 31, 2023}{Tallinn, Estonia}
\acmBooktitle{The 38th ACM/SIGAPP Symposium on Applied Computing (SAC '23), March 27-March 31, 2023, Tallinn, Estonia}
\acmDOI{10.1145/3555776.3577788}
\acmISBN{978-1-4503-9517-5/23/03}

\begin{document}
\title{A fully automated and scalable Parallel Data Augmentation for Low Resource Languages using Image and Text Analytics}

\renewcommand{\shorttitle}{Parallel Data Augmentation using Image and Text Analytics}

\author{Prawaal Sharma}
\affiliation{%
  \institution{Infosys}
  \streetaddress{Street Address}
  \city{Pune} 
  \state{Maharashtra, India} 
  \postcode{411057}
}
\email{prawaal_sharma@infosys.com}

\author{Navneet Goyal}
\affiliation{%
  \institution{BITS Pilani}
  \streetaddress{Street Address}
  \city{Pilani} 
  \state{Rajasthan, India} 
  \postcode{333031}
}
\email{goel@pilani.bits-pilani.ac.in}

\author{Poonam Goyal}
\affiliation{%
  \institution{BITS Pilani}
  \streetaddress{Street Address}
  \city{Pilani} 
  \state{Rajasthan, India} 
  \postcode{333031}
}
\email{poonam@pilani.bits-pilani.ac.in }

\author{Vishnupriyan K R}
\affiliation{%
  \institution{Infosys}
  \streetaddress{Street Address}
  \city{Chennai} 
  \state{Tamil Nadu, India} 
  \postcode{603002}
}
\email{vishnupriyan.r02@infosys.com}

\renewcommand{\shortauthors}{Prawaal et al.}

\begin{abstract}
Linguistic diversity across the world creates a disparity with the availability of good quality digital language resources thereby restricting the technological benefits to majority of human population. The lack or absence of data resources makes it difficult to perform NLP tasks for low-resource languages. This paper presents a novel scalable and fully automated methodology to extract bilingual parallel corpora from newspaper articles using image and text analytics. We validate our approach by building parallel data corpus for two different language combinations and demonstrate the value of this dataset through a downstream task of machine translation and improve over the current baseline by close to 3 BLEU points. 
\end{abstract}





\maketitle

\section{Introduction}

Modern NLP research focuses largely on the languages on the internet, which consists of only 20 of the 7,000 languages of the world \cite{only20}. This leaves the majority of languages understudied, which are also referred to as low-resource languages (LRLs), and are spoken by a large section of world population. LRLs can be described as \textit{resource scarce, under studied, less digitized, under privileged or less commonly taught}, among other denominations \cite{cieri}. There
are more than 2.5 billion inhabitants using 2,000 LRLs, within India and Africa and any progress for these languages shall help in digital enablement of these semi-literate populations \cite{LRL_Opp}.

Most NLP tasks (e.g. neural networks) require large amounts of training data. Availability of data therefore becomes more relevant in context of LRLs where scarcity of digital data is the primary challenge for taking NLP to masses. In this paper, we describe a novel approach using image and text analytics to build a completely automated, scalable and language agnostic methodology for bilingual parallel dataset generation.

\begin{figure}[t]
  \fbox{\includegraphics[width=0.95\linewidth]{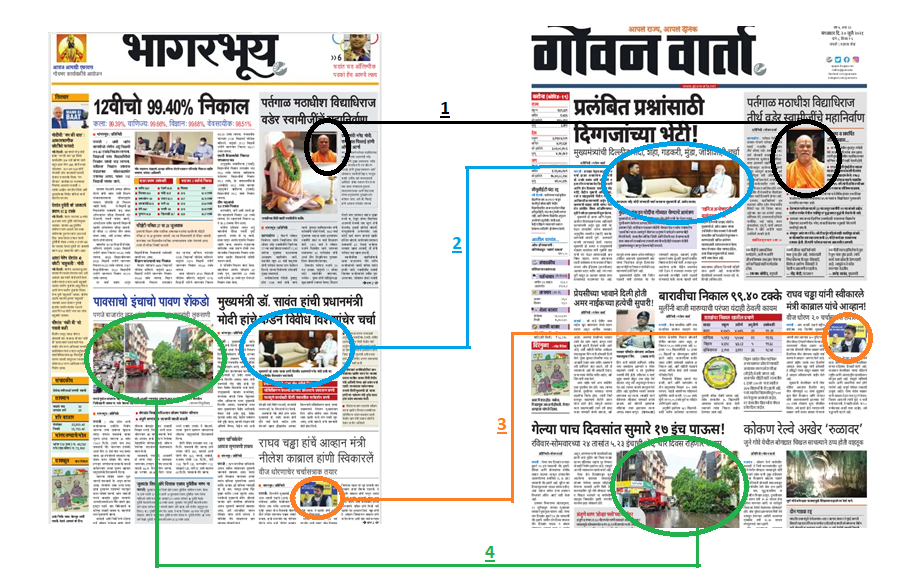}}
  \centering   
\addtolength{\belowcaptionskip}{-10pt}
\addtolength{\abovecaptionskip}{-5pt}

  \caption{Article mapping using images as pivots.}
  \label{Img_Matching}
\end{figure}

\begin{figure*}[t]
\centering
{
\fbox{\includegraphics[width=0.95\textwidth]{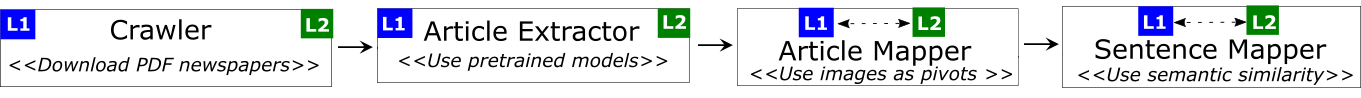}}
}
\addtolength{\belowcaptionskip}{-5pt}
\addtolength{\abovecaptionskip}{-5pt}
\caption{Proposed data augmentation pipeline}
\label{Method}
\end{figure*}

We use Konkani-Marathi as the primary example to establish our claims. Our choice for Konkani is based on the scarcity of digital resources along with its small population of native speakers \cite{LRLsurvey}. Marathi, on the other hand is resource rich and hence the combination would be instrumental in wide NLP applications. It has been established that the significance of parallel corpora is best observed with \textit{resource deficit-resource rich} language combination. 

As illustrated in Figure \ref{Img_Matching}, we observe that a lot of local newspapers in LRLs (published by large publishing houses) also have editions in other languages and they re-use the pictures across different language versions to optimise on resources. We apply this observation and use images as pivots for article mapping. Once articles are mapped, article text is extracted and sentences are mapped to form parallel corpus, which is then empirically evaluated. 

In a nutshell, the contributions from our work are three fold:

\begin{itemize}
  \item We use newspaper article images as pivots to map articles, which has not been explored earlier in similar context.
  
  \item We use language agnostic embeddings for sentence mapping (on LRL combinations) and empirically substantiate this.
  
  \item Our final Konkani-Marathi corpus is largest available dataset created without human annotations.
\end{itemize}

\section{Related Work}
Our work leverages the principles from image and text analytics to build sizeable parallel data corpus augmented from newspaper articles.

\subsection{Image Analytics}

The hypothesis for our work is that, news articles with same (or quasi same) images would have same information in text. We break this into primarily two parts (a) segmenting newspaper page into various article regions including pictures and (b) Matching images for article mapping.

Marking boundaries for articles, specially in a newspaper image (non machine readable) is a very difficult task. Most newspapers have multi column format with no explicit boundaries marked for individual articles. Existing article segmentation approaches include heuristic-based \cite{npextract}, graph embedding based \cite{graph} and deep learning based \cite{prima}. 

Image matching is based on feature detection of images. Feature detection is an abstraction of the image information and making a local decision at every image point to see if there is an image feature of the given type existing in that point. This should ideally be robust to image transformations such as rotation, scale, illumination, noise and affine transformations. The most popular image matching algorithms for this include scale invariant feature transform (SIFT), speed up robust feature (SURF) and binary robust independent elementary features (BRIEF). In the recent times, variants of neural networks including Convolutional Neural Network (CNN) has been explored and found more efficient than traditional image processing based techniques.

\subsection{Text Analytics}
For our work, the extracted and mapped articles need further processing on (a) text extraction and (b) sentence level mapping.

An Optical Character Recognition (OCR) system is a framework by which embedded textual information is repossessed through application of character extraction. Girdher et al. have done an extensive survey on Devanagari OCR and multiple approaches to accomplish this task \cite{girdher}. 

Sentence alignment across languages is an critical step for building bilingual (or multilingual) corpus. The existing approaches for sentence alignment are based on length based heuristics \cite{slas}, lexical correspondences \cite{salign3} and the recent deep learning based approaches which makes use of language agnostic sentence embedding (like LaBSE, which uses BERT like architecture), that can be compared using cosine-similarity \cite{labse}. 

\subsection{NLP for Konkani}

Existing NLP research for Konkani is limited to more elementary NLP tasks including POS tagging, sentiment analysis, NER etc. \cite{rajansentiment}. Indian Languages Corpora Initiative (ILCI) is a project of Technology Development in Indian Languages (TDIL), an agency of the Indian government. They have been engaged to create corpus for low resource Indian languages to facilitate research and avoid its extinction. They have created, two sets (across two different domains) of Hindi-Konkani corpus containing 25,000 sentences accomplished by human annotators \cite{rajansurvey}. The same dataset is used to accomplish the task of Neural Machine Translation (NMT) and a BLEU score of 23.5 has been achieved \cite{nmtkonkani}.  

\section{Methodology}
As illustrated in Figure \ref{Method}, the proposed data augmentation pipeline contains four components (a) Crawler, (b) Article Extractor, (c) Article Mapper and (d) Sentence Mapper. While \textit{crawler} and \textit{article extractor} work on the two languages (L1 and L2 as represented) independently, the \textit{mappers} (Article and Sentence) use the languages in conjunction and perform mappings.

\subsection{Crawler - Raw data collection}
\textbf{Crawler}, helps download copies of newspapers from online sources. Downloaded files are not machine readable and the content is “locked” in a snapshot-like image.  Crawler, helps in splitting the files into individual pages and label appropriately (with dates, page numbers and language code) to ensure easy referencing in the downstream processes.

\subsection{Article Extractor - Segmentation}
\textbf{Article Extractor}, helps with two functions, (a) Marking boundaries for individual articles (b) Extraction of images and text (using OCR) within the marked article. Embedded articles are considered parts of parent articles in our work due to its strong association with parent article. We use layout analysis dataset by Pattern Recognition and Image Analysis Research Lab (PRImA) for article boundary detection \cite{prima_A} and OpenCV \cite{opencv} to extract regions of interest (ROI) from segmented articles. For text extraction we consider a combination of EasyOCR, PaddleOCR and Tessaract and use majority voting for final decision.

\begin{figure}
\centering
{
\fbox{\includegraphics[width=0.4\textwidth]{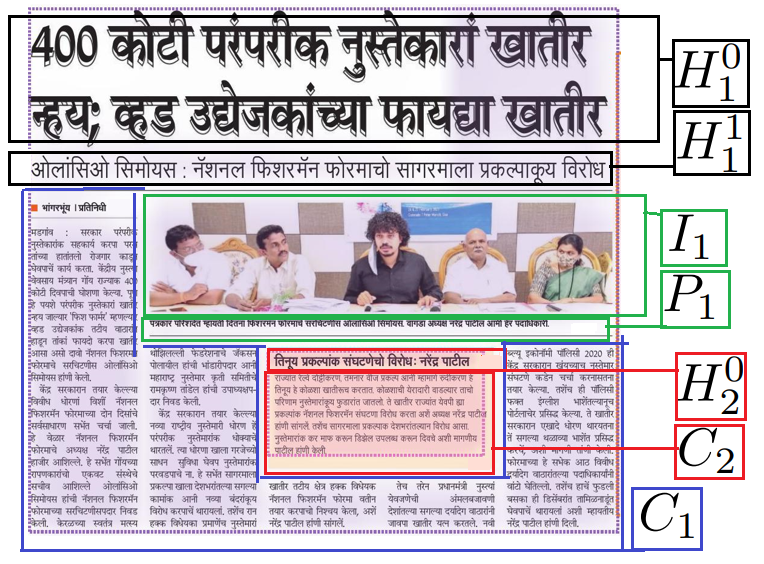}}
}
\addtolength{\belowcaptionskip}{-15pt}
\addtolength{\abovecaptionskip}{-5pt}
\caption{Article sub parts and nomenclature}
\label{Article Parts}
\end{figure}

As illustrated in Figure \ref{Article Parts} we partition the article into into four regions of interest (ROI) (a) Headlines ($H$) (which includes sub headlines) (b) Images ($I$) (c) Picture Captions ($P$) and (d) Contents ($C$). In case of embedded articles (e.g. $H_2^0$ and $C_2$ in Figure \ref{Article Parts}), indexing is done to maintain the hierarchy with parent article. 

\begin{equation}
a \equiv \begin{cases}
H_1^0,H_1^1,H_1^2... &\text{Main article headlines}\\
H_2^0,H_2^1,H_2^2... &\text{Embedded article headlines}\\
I_1,I_2,I_3...&\text{Images}\\
P_1,P_2,P_3...&\text{Picture captions}\\
C_1,C_2,C_3...&\text{Content}\\
\end{cases}
\end{equation}
\vspace{1pt}

Finally, each extracted article is stored as a set of image files and text file. Each ROI within the article is labelled with markers within the text file as illustrated in Equation 1.

\subsection{Article mapper - Match articles}
\textbf{Article Mapper}, compares the articles images ($I_i$,$I_j$) across the two languages ($L1,L2$) for similarity (for same date ($dt$)) and builds the set of mapped article tuples ($a_{i}^{L1},a_{i}^{L2}$) when image similarity score is beyond the set threshold. Embedded articles are mapped by comparing headline similarity. 
\begin{equation}
\{(a_{1}^{L1},a_{1}^{L2}),(a_{2}^{L1},a_{2}^{L2})...\} \equiv \operatorname*{\theta(}I^{L1}_i,I^{L2}_j)\\
\end{equation}
\begin{center}
\textit{$\theta$ is the image matching algorithm function}

$ I_i^{L1} \in \forall\ \text{\textit{(images for date (}dt\text{) \& language (}L1\text{))}} $

$ I_i^{L2} \in \forall\ \text{\textit{(images for date (}dt\text{) \& language (}L2\text{))}} $

\end{center}
\vspace{5pt}

We use SIFT as the image matching algorithm ($\theta$) for our work, since the valid image combinations are exact copies of each other and differ only in scale, illumination and shifts. 

\subsection{Sentence mapper - Match sentences}
The extraction and alignment of data till article level is straight forward task. However, the sentences between the mapped articles may not be positioned sequentially. Hence it becomes the most critical step in our experiment and impacts the overall accuracy of our experiment.

\textbf{Sentence Mapper}, helps to map sentences within the mapped articles $(a^{L1}_{i},a^{L2}_{i})$ for all sentences combinations applying semantic sentence similarity algorithms and build the set of mapped sentence tuples $(s_{j}^{L1},s_{j}^{L2})$. 
\begin{equation}
\{(s_{1}^{L1},s_{1}^{L2}),(s_{2}^{L1},s_{2}^{L2})...\} \equiv \operatorname*{\delta(}S^{L1}_k,S^{L2}_k)\\
\end{equation}
\begin{center}
\textit{$\delta$ is the semantic sentence similarity algorithm.}

\[
S_k^{L1} \in \forall\ \text{\textit{(sentences for article }} a^{L1}_{i}\text{\textit{)}} 
\]

\[
S_k^{L2} \in \forall\ \text{\textit{(sentences for article }} a^{L2}_{i}\text{\textit{)}} 
\]

\end{center}

We consider three types of metrics for sentence similarity ($\delta$), in our experiment (a) Language agnostic sentence embedding based cosine similarity (LAS), (b) Simple Length base heuristics (SLAS) and (c) Lexical overlap based metrics (LO). 

\textbf{Language Agnostic Sentence embeddings} converts sentence text into vectors to capture semantic information. These models are based on BERT-like architecture and is trained on 119 languages of different origins \cite{labse}. It claims to work universally including for the languages not part of its training corpus. We convert sentences into these vectors independently first and then find the cosine similarity between them, and refer as Language agnostic sentence embedding similarity (LAS) in our work.

\textbf{Sentence Length Alignment heuristics} is another metric to find the semantic similarity based on sentence length, and its position within the article. Sentence similarity is calculated as a function of number of words in a sentence, after filtering punctuation marks and applying adjustment factors for average sentence length across languages along with adjustment for the size of articles mapped \cite{slas}.

\textbf{Lexical Overlap} is based on precision, recall and F-Score on common words across the candidate sentence set to measure the overlap statistically using a pivot language. We use English as a pivot language and perform lexical translation to English for both the languages. We use google translate for Marathi to English lexical conversion and English-Konkani dictionary by Maffei and Xavier \cite{konkanidict} for Konkani to English lexical conversion.

\section{Experimental Setup and Results}

We run our experiment in two parts, (a) Intrinsic evaluation using multiple language combinations, (b) Extrinsic evaluation using downstream task of machine translation (MT) . We use Konkani-Marathi as the primary language combination to study multiple sentence mapping methods, and later apply the same process on Punjabi-Hindi language pair to establish that our methodology is language agnostic and works universally. 

\begin{table*}[t]
  \begin{minipage}{.58\linewidth}
    \centering
    \begin{tabular}{l|c|c|c|c|c|c}
      \toprule       
       &\multicolumn{3}{c|}{Sentence Lengths} & \multicolumn{3}{c}{Article Lengths} \\       
       Mapping & (1-10) & (11-19) & 20+ & (1-5) & (6-15) & (16+)\\
         Strategy  & words & words & words & sentences & sentences & sentences\\
      \midrule
        LAS  & 3.8 & 3.7 & 3.8 & 3.8 & 3.8 & 3.7\\
        SLAS & 3.4 & 3.4 & 3.2 &  3.1 & 3.5 & 3.3\\
        LO & 2.9 & 3.0 & 2.6 &  2.8 & 2.9 & 2.9 \\
      \bottomrule
    \end{tabular}
    \caption{STS for various sentence and article size}\label{tab:first}
  \end{minipage}%
  \begin{minipage}{.4\linewidth}
    \centering
    \begin{tabular}{l|c|c}
      \toprule          
        \multirow{2}{*}{Variables} & \multicolumn{2}{c}{Value}\\
         & Konkani-Marathi & Punjabi-Hindi\\          
      \midrule
        {Mapped articles}  & 1,320 & 150\\
        {Mapped sentences}  & 14,448 & 2,200\\
        {Human evaluation}  & 600 & 100\\
        {STS Score}  & 3.70 & 3.73\\
      \bottomrule
    \end{tabular}
    \caption{Summary view under various heads}\label{tab:second}
  \end{minipage}
\end{table*}

\subsection{Konkani-Marathi parallel corpus evaluation}
We apply our methodology (as described in Section 3) and build a parallel corpus for Konkani-Marathi using all sentence mapping methods independently. We analyse the quality of each corpus by applying human annotation on a subset of sentence pairs from our experiment. For defining the annotation scores, we use Semantic Textual Similarity (STS), characterised by six ordinal levels ranging from complete semantic equivalence (5) to complete semantic dissimilarity (0) \cite{semval}. This is the most widely used evaluation metric for parallel data augmentation tasks, and used by multiple people working on this field. 

We have sampled a total of 900 sentences, in two phases. In the first phase we have sampled 200 pairs from each sentence alignment strategy (making a total of 600 sentence pairs) and analysed the results. In the second phase we have sampled, another 300 pairs for the most appropriate sentence mapping strategy as evident from first phase analysis. The sampling was stratified, shuffled randomly such that no ordering is preserved.  We illustrate the evaluation in two parts as below.

\textbf{Phase 1:} As illustrated in Table \ref{tab:first} the human evaluation  for different sentence and article sizes clearly indicate that LAS gives the best results and is most appropriate for our task. 

\textbf{Phase 2:} Based on the results from first phase, we use LAS as our final alignment method. As illustrated in Table \ref{tab:second} our Konkani-Marathi parallel corpus contains 14,448 sentence-pairs (aggregating a total of 28,896 sentences extracted). We evaluate STS on a corpus of 500 sentences (200 from first phase LAS evaluation and another 300 in second phase) and observe that the average STS in our corpus is 3.7 and more than 92\% of our mapped sentences, have STS score of greater than 3.

\begin{figure}[b]
\centering
{
\fbox{\includegraphics[width=0.4\textwidth]{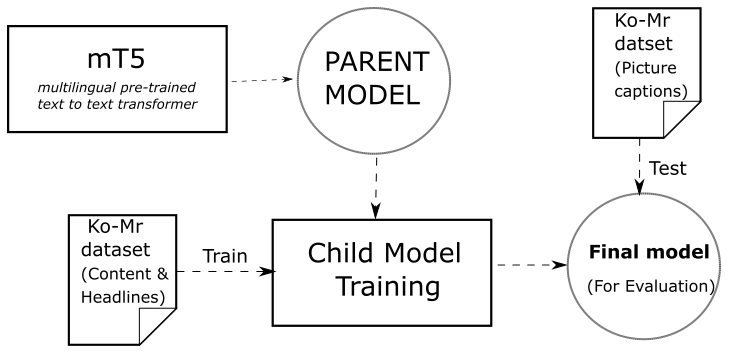}}
}
\caption{Proposed model for our evaluation}
\label{Experiment}
\end{figure}

\subsection{Punjabi-Hindi evaluation}
We execute the same experiment on Punjabi-Hindi language combination, following the exact same steps on a smaller scale. Our intent of doing this as an extended experiment is to analyse and validate the significance of our work on other language pairs. As illustrated in Table \ref{tab:second} the results on this language pair is at par with our primary experiment. 

\subsection{Case study: MT for Konkani-Marathi}
To access the quality of the entire Konkani-Marathi aligned corpus, we perform extrinsic evaluation using the downstream machine translation (MT) task which shall leverage the entire aligned corpus as an input. 

As illustrated in Figure \ref{Experiment}, we use mT5 (multilingual pre-trained text to text transformer) as our parent model \cite{mt5}, and further fine tune with our Konkani-Marathi parallel corpus (headlines and article content). We have used picture captions as ground truth to test our translation model and achieved the BLEU score of 26.4 which is an improvement over the current baseline for Konkani, by around 3 BLEU points (compared to existing baseline of 23.5 as mentioned in Section 2.3). This further validates the overall quality of the dataset. 

\section{Conclusion and Future Work}
Our results indicate that our methodology is agnostic of language combinations and reasonably good to build a scalable model for parallel corpus generation\footnote{https://github.com/prawaal/Konkani-Marathi-Data-Corpus/}. We conclude that in order to boost the quality of the dataset additional constraints may need to be provided. The scale of the dataset can be enhanced by considering images clicked by different people for the same news event. We aim to explore these in our future work.

\bibliographystyle{ACM-Reference-Format}
\bibliography{sample-bibliography} 

\end{document}